\documentclass[12pt]{article}

\usepackage{fullpage}
\usepackage{setspace}
\usepackage{parskip}
\usepackage{titlesec}
\usepackage{xcolor}
\usepackage{lineno}

\usepackage{times}

\PassOptionsToPackage{hyphens}{url}
\usepackage[colorlinks = true,
            linkcolor = blue,
            urlcolor  = blue,
            citecolor = blue,
            anchorcolor = blue]{hyperref}
\usepackage{etoolbox}
\makeatletter
\patchcmd\@combinedblfloats{\box\@outputbox}{\unvbox\@outputbox}{}{%
  \errmessage{\noexpand\@combinedblfloats could not be patched}%
}%
\makeatother

\usepackage{natbib}

\renewenvironment{abstract}
  {{\bfseries\noindent{\abstractname}\par\nobreak}\footnotesize}
  {\bigskip}

\titlespacing{\section}{0pt}{*3}{*1}
\titlespacing{\subsection}{0pt}{*2}{*0.5}
\titlespacing{\subsubsection}{0pt}{*1.5}{0pt}

\usepackage{authblk}

\usepackage{graphicx}
\usepackage[space]{grffile}
\usepackage{latexsym}
\usepackage{textcomp}
\usepackage{longtable}
\usepackage{tabulary}
\usepackage{booktabs,array,multirow}
\usepackage{amsfonts,amsmath,amssymb}
\providecommand\citet{\cite}
\providecommand\citep{\cite}

\newif\iflatexml\latexmlfalse

\AtBeginDocument{\DeclareGraphicsExtensions{.pdf,.PDF,.eps,.EPS,.png,.PNG,.tif,.TIF,.jpg,.JPG,.jpeg,.JPEG}}

\usepackage[utf8]{inputenc}
\usepackage[english]{babel}

\begin{document}

\title{Robustness to fundamental uncertainty in AGI alignment}

\author[1]{G Gordon Worley III}
\affil[1]{Phenomenological AI Safety Research Institute}

\date{\today}

\begingroup
\let\center\flushleft
\let\endcenter\endflushleft
\maketitle
\endgroup

\selectlanguage{english}
\begin{abstract}
The AGI alignment problem has a bimodal distribution of outcomes with most outcomes clustering around the poles of total success and existential, catastrophic failure. Consequently, attempts to solve AGI alignment should, all else equal, prefer false negatives (ignoring research programs that would have been successful) to false positives (pursuing research programs that will unexpectedly fail). Thus, we propose adopting a policy of responding to points of philosophical and practical uncertainty associated with the alignment problem by limiting and choosing necessary assumptions to reduce the risk of false positives. Herein we explore in detail two relevant points of uncertainty that AGI alignment research hinges on---meta-ethical uncertainty and uncertainty about mental phenomena---and show how to reduce false positives in response to them.
\end{abstract}

\section*{Introduction}

The development of artificial intelligence (AI) is making possible many technological advancements and improvements to well-being but also exposes humans to considerable risks \citep{al2018}. Key among these risks are the existential risks associated with artificial general intelligence or AGI \citep{Turchin_2018}. In particular, it is likely that the development of superintelligent AGI will create either existential catastrophe or massive benefits for humanity with little room for mildly bad, mildly good, or neutral outcomes \citep{bostrom2014}. This bimodal distribution of outcomes where one of the outcomes is extremely undesirable implies that we are better off if we focus more on avoiding negative outcomes and less on achieving positive outcomes, even at the risk of missing out on much possible positive value \citep{bostrom2003}. Thus, assuming AGI will be built eventually, we can increase the expected value of AGI by working on interventions that reduce the chance of existential catastrophe, and we call the field that focuses on these interventions AI safety \citep{yampolskiy2012}.

Central among interventions that address AI safety is alignment, the problem of how to build AGI that is aligned with human interests \citep{2017}. The alignment problem is complex and likely requires solving many open problems in mathematics, economics, computer science, and philosophy \citep{yudkowsky2016}. This is unfortunate from the perspective of increasing the expected value of AGI by reducing existential risks because it introduces many opportunities for researchers to make mistakes. If those mistakes are false negatives, like believing an alignment scheme won't work when it would, then there is a comparatively small loss of value from failing to develop safe AGI as soon as possible, but if those mistakes are false positives, like believing an alignment scheme will work when it won't, then there is an astronomically large loss of value from accidentally developing unsafe AGI. This implies that, all else equal, we are better off preferring false negatives to false positives when working on alignment.

This preference implies several AGI research and development policies. One previously explored by Yudkowsky is the idea of security mindset, borrowing an idea of the same name from computer security researcher Bruce Schneier, which we might summarize as the expectation of operating in an adversarial environment that is actively trying to produce unsafe AGI \citep{yudkowsky2017}. Another, which we will demonstrate the value of here, might be called robustness to fundamental uncertainty, the idea that we can reduce false positives by choosing necessary assumptions when reasoning about alignment such that the chosen assumptions minimize the risk of false positives among the space of alternative assumptions. We'll show how to reason about risks from fundamental uncertainty by examining two cases relevant to AI alignment---meta-ethical uncertainty and uncertainty about mental phenomena---and recommend specific assumptions to reduce the risk of existential catastrophe from AGI.

\section*{Reducing false positive risk}

If we wish to reduce false positives in AI alignment research, viz. reduce the chance of accidentally producing unaligned AGI when we thought we would have produced aligned AGI, we can work to reduce the overall rate of errors or we can trade-off producing more false negatives against producing fewer false positives \citep{Neyman_1933}. The former is hard to control since it largely depends on the base error rates of individual researchers, but the the trade-off between false negatives and false positives may be adjusted by taking actions that prefer one to the other. For example, to decrease false negatives and increase false positives researchers might try more ideas they think less likely to work and vice versa to increase false negatives and decrease false positives. The latter is the sort of approach to reducing false positives one might get in part by application of Yudkowsky's ideas around security mindset.

Another approach to reducing the rate of false positives by trading off against false negatives is to apply the principle of parsimony (also known as Occam's razor) to decrease the probability of any particular attempt at producing aligned AGI failing by reducing the number of variables multiplied together in calculating the probability of success \citep{pearl2000}. On the one hand this suggests the adoption of the well-known engineering principle that simpler systems are less likely to fail, and on the other suggests we can reduce false positives by making a deep commitment to reducing the number of variables, especially implicit variables that are often ignored, when building aligned AGI \citep{allan1992}. We cannot, however, remove all variables from our reasoning. One cause of our need to reason with irreducible variables is computational because complete deductive proofs cannot always be found within reasonable time bounds \citep{Leike_2015}. The other, more pernicious cause of this need is the problem of epistemic circularity.

Briefly stated, epistemic circularity is the problem that nothing can be reliably known without first knowing something reliably \citep{Alston_1986}. There have been numerous attempts to solve epistemic circularity since it was first identified by Pyrrhonian skeptics in ancient Greece via the Problem of the Criterion, and it remains a key problem in epistemology today because it likely has no complete solution \citep{Lammenranta_2003}. Instead we are left to adopt the least-bad pragmatic solution, known as particularism, by choosing to assume unreliable entitlements to some knowledge and then reasoning as if we knew those facts reliably \citep{chisholm1973}, \citep{alston1993}, \citep{alston2005}. This does not fundamentally address the skepticism that epistemic circularity implies, but it does allow us to contain our skepticism to only a few philosophical assumptions---called hinge propositions by Wittgenstein, analogous to axioms in formal logics, and related to approximation of the universal prior in Solomonoff induction---to get on with reasoning in spite of epistemic circularity \citep{talvinen2009}.

Adopting particularism still leaves us with the problem of choosing the specific assumptions upon which we will build our reasoning about how to build aligned AGI. Since these choices cannot be made reliably they stand as ``free'' variable in our reasoning that we must choose using criteria other than likelihood of being true since that likelihood cannot be adequately assessed. This need to choose gives us an opening, though, to choose such that the risk of false positives is reduced by considering the relative likelihood of false positives given different choices of assumptions. We will now review some of these assumptions, why they are necessary, and evaluate their expected impacts on the likelihood of false positives in order to conclude what assumptions are safest to make.

\section*{Necessary assumptions}

We have identified at least two problems, both philosophical and pragmatic, that presently require making assumptions when reasoning about AGI alignment: meta-ethical uncertainty and uncertainty about mental phenomena. For each problem we consider why making an assumption is necessary and give arguments in favor of and against particular assumptions in terms of their impact on false positives and likelihood of success in order to determine the safest assumption to make.

\subsection*{Meta-ethical uncertainty}

AGI alignment is typically phrased in terms of aligning AGI with human interests, but this hides some of the complexity of the problem behind determining what ``human interests'' are. Taking ``interests'' as a synonym for the cluster of concepts we also call ``values'' and ``preferences'', we can begin to make some progress by treating alignment as at least partially the problem of teaching AGI to learn human values \citep{soares2016}, \citep{scheutz2017}. Unfortunately, what constitutes human values is currently unclear since humans may not be aware of the extent of their own values, human values may be partial, and humans may not hold reflexively consistent or rational values \citep{Scanlon2003-SCARO}, \citep{pettit1991}, \citep{Tversky_1969}. This creates a problem in terms of alignment because the values of individual humans, let alone the combined values of humanity, contradict each other, so in order for an AGI to align its behavior with human values it must have some way of resolving those conflicts since an AGI is very likely to be modelable as having a utility function and thus the AGI must meet rationality constraints requiring strong consistency of preferences \citep{yudkowsky2004}. Additionally, it appears that AGI cannot learn human values without making at least some normative assumptions as a result of a no free lunch theorem in the theory of value learning that requires assuming norms in order to enable extracting values from observed human behavior \citep{Armstrong2017OccamsRI}. Consequently this means alignment requires making normative assumptions to resolve value conflicts, and doing so means alignment asks us to tackle the ethical issue of normative uncertainty \citep{macaskill2014}.

Normative uncertainty is a symptom of deeper uncertainty in ethics caused by meta-ethical uncertainty about the existence of moral facts because epistemic circularity prevents us from reliably knowing whether or not moral facts exist, let alone what moral propositions are true if moral facts exist \citep{chisholm1982}. Meta-ethical uncertainty forces us to speculate about moral facts because knowledge about moral facts, even if it is the knowledge that moral facts do not exist, is necessary to ground ethical reasoning \citep{zimmerman2010}. Thus in order to reason about alignment we must consider what hinge proposition to adopt about the existence of moral facts in order to be able to design AGI that can behave in a manner aligned with conflicting human values and indeed successfully learn human values at all. The standard positions regarding the existence of moral facts are realism, anti-realism, and skepticism being respectively for, against, and uncertain about their existence, so we'll consider their effects on false positives in AGI alignment in turn.

\subsubsection*{Moral realism}

If we suppose realism, then we could build aligned AGI on the presupposition that it could at least discover moral facts even if no moral facts were specified in advance and then use knowledge of these facts to resolve conflicts in human values and learn human values. Now suppose this assumption is false and that moral facts do not exist or even if they do exist they cannot be known, then our moral-facts-assuming AGI would either never discover any moral facts to guide its behavior or would assume arbitrary moral propositions to be facts that would not be sure to produce human-preferred resolutions to value conflicts or enable learning human values successfully. Such an AGI might still achieve de facto alignment with human values if it adopted moral propositions it believed to be facts that allowed it to converge on a functionally equivalent solution to the one an AGI that had correctly assumed no knowledge of moral facts would have used, but lacking an argument suggesting such an AGI would be less likely to result in false positives than a simpler AGI that started out not assuming knowledge of moral facts, this seems a strictly more risky approach.

To make this argument about the risks of false positives concrete, consider efforts to align a realist AGI in a world where moral facts do not exist. Let us suppose, for the sake of argument, that the AGI believes "do unto others as you would have them do unto you" or the Golden Rule is a moral fact contrary to the reality of this world. Believing the Golden Rule to be true, the AGI resolves conflicts between the values of individual humans on its basis, possibly contrary to their preferences.

For example, Alice may prefer that she and only she eats cake, and Bob may prefer that he and only he eats pie. Simultaneously, Alice prefers to eat cake to pie, and Bob prefers to eat pie to cake. This seems fine, since their combined preferences should allow Alice to eat cake and Bob to eat pie, but by the Golden Rule the AGI may reason Alice does not want to eat cake since she prefers others not to eat cake and Bob does not want to eat pie since he prefers others not to eat pie. That is, reasoning generally that humans may have partial, inconsistent preferences, the AGI attempts to extract their complete, consistent preferences on the basis of making them fit the moral fact of the Golden Rule. This results in a strictly worse outcome for Alice and Bob than if the AGI had not attempted to align itself with their preferences on the assumption of this particular moral fact given they live in a world without moral facts.

\subsubsection*{Moral anti-realism}

If we suppose anti-realism, then we must build aligned AGI to reason about conflicts in human values and to learn human values in the absence of any normative assumptions grounded in moral facts. Now suppose this assumption is false and moral facts do exist, then our moral-facts-denying AGI would resolve value conflicts and learn human values in a way not based on moral facts and would fail to act in a way that fully satisfied human preferences. Such an AGI might still achieve de facto alignment with human values if it adopted conflict resolution mechanisms and value learning norms that were functionally equivalent to those it would have adopted if it had known moral facts, and it may be able to do this because the unacknowledged moral facts would impact such an AGI through their influence on human values. Although less efficient than acting on moral facts directly, this suggests an anti-realist AGI could still stand a chance of aligning itself based on moral facts as revealed by the human values being aligned with in a world where moral facts exist and be less likely to suffer from false positives than a realist AGI.

To make the argument clear, like in the previous section, let's consider the case where we build an AGI, although this time it is anti-realist and it exists in a world where moral facts actually exist. Specifically, for the sake of argument, let's suppose the Golden Rule is a moral fact in this world. Not believing in moral facts, though, the AGI looks to other sources for norms to resolve value conflicts and to extract human values from observations of behavior.

For example, Alice and Bob may have their preferences as in the previous example, and also a shared preference for fairness. Supposing that Alice and Bob are the sum total of all humans in this world, the AGI learns that all humans prefer fairness and so adopts it as a norm (assuming the AGI has been programmed with a meta-norm of adopting those values that all humans express). Enter into this situation a day when Alice would like to eat pie and Bob would like to eat cake. The AGI can now act to assist or hinder Alice and Bob in one of two ways consistent with their preference for fairness: either neither is allowed to eat cake or pie or both are allowed to eat cake and pie.

If the AGI acts to prohibit eating either, it will coincidentally act in accord with the Golden Rule. If the AGI acts to allow eating either, it will coincidentally violate the Golden Rule. In the former case Alice and Bob will be, perhaps even unbeknownst to them, getting more of what they value because it satisfies the moral fact of their world of the Golden Rule, and in the latter case will be getting less of what they want, also possibly without their explicit knowledge. Unknown to the AGI and possibly unknown to Alice and Bob, fairness is actually a proxy value for the Golden Rule, but because the AGI adopts fairness as a norm rather than the Golden Rule directly it may, on the margin in situations like the one described, act in ways contradictory to the Golden Rule, thus rendering the anti-realist AGI less efficient at satisfying human values than a realist AGI in this world where moral facts exist.

\subsubsection*{Moral skepticism}

If we suppose skepticism, then we must build aligned AGI in the absence of any knowledge of moral facts, which is similar to the anti-realist case, but different in that there we assume moral facts do not exist where here we remain open to the possibility that moral facts may exist even if we don't or can't know them. Now suppose this assumption is false and we can know about the existence of moral facts such that we could, even if only in theory, decide in favor of realism or anti-realism, then the AGI may fail to acknowledge and use this knowledge to make more informed decisions about value conflict resolution and to better learn human values, however we could reasonably expect this to be mitigated because skepticism, unlike realism and anti-realism, need not assume a strong metaphysical claim and only instead claim that perfect knowledge is not possible, thus the skeptical AGI could switch to believing realism or anti-realism with high credence and act on that belief. This still leaves open the practical question of how a skeptical AGI would address value conflicts and choose norms for learning human values, but it could reasonably either choose norms the same way an anti-realist AGI would until it learned more, or it could adopt norm particularism to ground its norms on moral assumptions after the style of Dancy's ethical particularism despite not being able to make absolute truth claims about which norms are best \citep{DANCY_1983}. However it chooses norms, in having greater flexibility of choice about moral facts it mitigates the risk of false positives from the assumption it makes about their existence.

In this case our example will come in two parts. For both, let's suppose we construct a morally skeptical AGI in a world where either moral facts definitely exist or don't. Again, we will specifically consider the case of the moral fact of the Golden Rule existing or not.

If our skeptical AGI finds itself in the world where moral facts exist, it faces a similar situation to the anti-realist AGI in the previous section. However, unlike the anti-realist AGI, because our skeptical AGI has not decided against moral facts, it can provisionally adopt the belief that moral facts do exist and search for them. Depending on the world it may or may not be able to discover that the Golden Rule is a specific moral fact that holds in it, but regardless of this it can at least move in the direction of more accurately acting in accordance with the world it finds itself in. It could then, for example, move towards an adoption of norms that might seemingly violate the observed preferences of Alice and Bob but would better satisfy their values given that it would include in this world the moral fact of the Golden Rule. In practice this might look like first adopting the fairness norm the way the anti-realist AGI did, then learning more and updating to the Golden Rule and applying it more strongly, thus eliminating the efficiency losses the anti-realist AGI suffered in the realist world.

If our skeptical AGI finds itself in a world where moral facts do not exist, it's in a similar but much better situation than the realist AGI in the second previous section. Unlike the realist AGI, the skeptical AGI need not assume knowledge of any specific moral fact, like the Golden Rule, or even believe it can discover any moral facts, so it can instead look to discover norms from elsewhere. Rather than adopting the Golden Rule, it might adopt a fairness norm assuming Alice and Bob both express this preference and it could use this to resolve value conflicts between them and infer values from their behavior. The adoption of this norm would remain contingent on observations, though, and not be held as fact, so it would not risk acting against human values by thinking it knew human values better than the human did by virtue of knowledge of non-existent moral facts.

\subsubsection*{Results}

The result of this analysis is that it is best to build AGI assuming moral skepticism to reduce the risk of false positives when building aligned AGI. Although it may be a much less direct route to aligned AGI than assuming realism would be if realism were true and a slightly less direct route than assuming anti-realism if anti-realism were true, skepticism allows AGI safety research to avoid committing to a hinge proposition that will much raise the risk of false positives and avoids unnecessarily hindering AGI efficiency at human value satisfaction.

It's worthwhile to note that we don't necessarily need to build AGI that are aligned to ``true'' values or moral facts if they contradict human preferences. It is only necessary to deal with normative uncertainty because of conflicts that arise in aggregating values and difficulties in inferring values from observed behavior. Absent these concerns it would not matter what hinge propositions, if any, were adopted concerning meta-ethical uncertainty. Similarly we only consider the realist/anti-realist divide because it is a clear hinge on which all norm selection choices implicitly depend, and analysis of other, more modern, complex, and nuanced ethical and moral positions may be worthwhile if a choice about them must necessarily be made. Doing a similar analysis of other hinge propositions necessitated by meta-ethical uncertainty is a potentially valuable direction for future research, as is evaluating particular norms that might be adopted by an AGI in the same way we are here evaluating hinge propositions to minimize the likelihood of false positives and minimize existential risk.

\subsection*{Uncertainty about mental phenomena}

Schemes to align AGI with human values depend on the dispositions of AGI, and one disposition AGI has is to subjective experience, consciousness, and mental phenomena \citep{Adeofe}, \citep{Nagel_1974}. Whether or not we expect AGI to realize this disposition matters because it influences the types of alignment schemes that can be considered since an AGI without a mental aspect can only be influenced by modifying its algorithms and manipulating its behavior whereas an AGI with a mind (and for our purposes here, is ``conscious'') can be influenced by engaging with its perceptions and understanding of the world \citep{dreyfus1978}. In other words we might say mindless AGI can be aligned only by algorithmic and behavioral methods whereas conscious AGI can also be aligned by ``philosophical'' methods that work on its epistemology, ontology, and axiology \citep{brentano1995}. It's unclear what we should expect about the mentality of future AGI, though, because we are presently uncertain about mental phenomena in general (cf. the work of Chalmers and Searle for modern, popular, and opposing views on the topic), so we are forced to speculate about mental phenomena in AGI when we reason about alignment \citep{chalmers1996}, \citep{searle1984}.

Note, though, that this uncertainty may not be fundamental \citep{dennett1991}. For example, if materialist or functionalist attempts to explain mental phenomena prove adequate, perhaps because they lead to the development of conscious AGI, then we may agree on what mental phenomena are and how they work \citep{Oizumi_2014}. If they don't, though, we'll likely be left with metaphysical uncertainty around mental phenomena that's rooted in the epistemic limitations of perception \citep{hussrl2014}. Regardless of how uncertainty about mental phenomena might later be resolved, it currently creates a need for pragmatically making assumptions about it in our reasoning about alignment. In particular we want to know whether or not we should design alignment schemes that assume a mind, even if we expect mental phenomena to be reducible to other phenomena. Given that we remain uncertain and cannot dismiss the possibility of conscious AGI, what we decide depends on how likely alignment schemes are to succeed and to avoid false positives conditional on AGI having the capacity for mental phenomena. The choice is then between whether we design alignment schemes that work without reference to mind or whether they engage with it.

\subsubsection*{Initial false positive analysis}

If we suppose AGI do not have minds, whether because we believe they have none, are inaccessible to us, or not causally relevant to alignment, then alignment schemes can only address the algorithms and behavior of AGI. Such schemes include but are not limited to approaches that aim to directly encode values in AGI or focus on inverse reinforcement learning to discover human values \citep{HadfieldMenell2016CooperativeIR}. Now suppose this assumption is false and AGI do have minds, then our alignment schemes that work only on algorithms and behavior would be expected to continue to work since they function without regard to the mental phenomena of AGI, making the minds of AGI irrelevant to alignment. This suggests there is little risk of false positives from supposing AGI do not have minds.

If we suppose AGI do have minds, then alignment schemes can also use philosophical methods to address the values, goals, models, and behaviors of AGI. Such schemes would likely take the form of ensuring that updates to an AGI's ontology and axiology converge on and maintain alignment with human interests \citep{blanc2011}, \citep{armstrong2015}. Now suppose this assumption is false and AGI do not have minds, then our alignment schemes that employ philosophical methods will likely fail because they are attempting to address mechanisms of action not present in AGI. This suggests there is a risk of false positives from supposing AGI have minds proportionate with the likelihood that we do not build conscious AGI.

From this analysis it seems we should suppose mindless AGI when designing alignment schemes so as to reduce the risk of false positives, but note that we did not consider the likelihood of success at aligning AGI using only algorithmic and behavioral methods. That is, all else may not be equal between these two assumptions such that the one with the lower risk of false positives might not be the better choice if we have additional information that leads us to believe that alignment of conscious AGI is much more likely to succeed than the alignment of mindless AGI, and it appears that we have such information in the form of Goodhart's curse.

\subsubsection*{Revised analysis in light of Goodhart's curse}

Goodhart's curse says that when optimizing for the measure of a target value the optimization process will implicitly maximize divergence of the measure from the target value \citep{yudkowsky2017a}. This is an observation that follows from the combination of Goodhart's law and the optimizer's curse.

As originally formulated, Goodhart's law says ``Any observed statistical regularity will tend to collapse once pressure is placed upon it for control purposes'' \citep{Goodhart_1984}. A more accessible expression of Goodhart's law, though, would be that when a measure of success becomes the target, it ceases to be a good measure \citep{strathern1997}. A well known example of Goodhart's law comes from a program to exterminate rats in French-colonial Hanoi, Vietnam: the program paid a bounty for rat tails on the assumption that a rat tail represented a dead rat, but rat catchers would instead catch rats, cut off their tails, and release the rats so they could breed and produce new rats so their tails could be turned in for more bounties \citep{vann2003}. There was a similar case with bounties for dead cobras in British-colonial India that intended to incentivize the reduction of cobra populations that instead resulted in the creation of cobra farms \citep{siebert2001}. Additional examples abound: targeting easily-measured clicks rather than conversions in online advertising, optimizing for profits over company health in business, and unintentionally incentivizing publication count over academic progress in academia \citep{shraga2014}, \citep{jackall2009}, \citep{nature2010}.

The optimizer's curse observes that when choosing among several possibilities, if we choose the option that is expected to maximize value, we will be ``disappointed'' (realize less than the expected value) more often than average \citep{harrison1984}. This happens because optimization acts as a source of bias in favor of overestimation, even if the estimated value of each option is not biased itself \citep{Smith_2006}. For example, a company trying to pick a project to invest in to earn the highest rate of return will consistently earn less return than predicted due to the optimizer's curse \citep{chen2009}. Similarly a person trying to pick the best vacation will, on average, have a worse vacation than expected because the vacation that looks the best is more likely than the other options to look that way due to overestimation. Further, the effect of the optimizer's curse is robust: mitigating it is theoretically possible but requires certainty about probability distributions that is impossible to obtain in practice and even if an agent satisfices (accepts the option with the least expected value that is greater than neutral) rather than optimizes they will still suffer more disappointment than gratification \citep{Smith_2006}, \citep{marks2008}. Consequently, when combined with Goodhart's law, the effect is that attempts to optimize, however weakly, for a measure of success result in increased likelihood of failure to hit the desired target---Goodhart's curse.

This tendency of measure and target value to diverge under optimization is also known as ``regressional Goodharting'' and is a special case of a more general phenomenon known simply as ``Goodharting'' \citep{garrabrant2018}. We need not especially concern ourselves with these more general cases, though, because Goodhart's curse is enough to make us reconsider the likelihood of successfully aligning AGI with human value by attempting to optimize their algorithms and behaviors rather than engaging with their minds.

Suppose we have an alignment scheme that proposes to align an AGI with human values via humans manipulating its algorithm or the AGI observing human behavior to infer human values. In the case of manipulating its algorithm, humans are preferentially choosing (viz. optimizing for) changes that result in better expectation of alignment. In the case of the AGI observing human behavior, it is trying to determine the best model of human values that explains the observed behavior. Both cases create conditions for alignment of the AGI to suffer from Goodhart's curse and therefore diverge from actual human values by optimizing for something other than human values themselves, namely appearance of alignment as observed by humans and strength of model fit as measured by the AGI.

This argument implies that attempts to align mindless AGI via algorithmic and behavioral methods will reliably result in AGI that diverge from human values and thus fail of align with human interests. This leads us to conclude that, although assuming conscious AGI has a greater risk of false positives than assuming mindless AGI all else equal, all else is not equal, mindless AGI is unlikely to be capable of maintaining alignment, and thus we are forced to take on the risks associated with assuming conscious AGI when designing alignment schemes due to the low probability of success from assuming otherwise.

\section*{Conclusion}

As argued, since the development of AGI has a bimodal distribution of outcomes that causes us to favor false negatives to false positives to better avoid catastrophe, we are best off if we choose epistemically necessary hinge propositions (assumptions) that minimize false positives all else equal. We considered two such cases where assumptions are necessary.

In the first case, because AGI alignment requires making assumptions about norms over human values, alignment necessitates resolving questions of meta-ethical uncertainty which means, in part, choosing an assumption about the existence of moral facts. We evaluated the possible assumptions of moral realism, anti-realism, and skepticism, and found realism to have the highest risk of false positives and skepticism to have the least. For that reason we recommend AGI safety researchers assume and design AGI to assume moral skepticism.

In the second case, because AGI alignment schemes may depend or not on AGI having minds, we considered the hinge proposition that AGI will be conscious. Although it turned out that, in terms of risk of false positives, it was safer to assume AGI will be mindless when designing an alignment scheme, because of Goodhart's curse it is substantially unlikely that alignment of mindless AGI is possible. Consequently we recommend AGI researchers to design alignment schemes that assume conscious AGI.

These are not necessarily all the assumptions that must be made in the construction of AGI aligned with human values, but they are at least two that must be, and additional evidence and arguments may lead us to conclude that a different bundle of assumptions are better responses to the philosophical and practical points of uncertainty that forced us to make assumptions. Future research should consider additional points of uncertainty that demand we make assumptions about AGI alignment to better understand the distribution of alignment schemes that are likely to result in acceptable outcomes for humanity.

\selectlanguage{english}
\clearpage
\bibliographystyle{aaai-named}
\bibliography{robust.bib}

\end{document}